\pdfoutput=1

\documentclass[11pt]{article}

\usepackage{emnlp2021}

\usepackage{times}
\usepackage{latexsym}
\usepackage{graphicx} 
\usepackage{booktabs} 
\usepackage{placeins} 
\usepackage{microtype} 
\usepackage{enumitem} 
\usepackage{amsmath, bm} 
\usepackage[skip=2pt]{subcaption} 
\usepackage{comment} 
\usepackage[labelfont=bf]{caption} 
\newenvironment{packed_enum}{
\begin{enumerate}
  \setlength{\itemsep}{1pt}
  \setlength{\parskip}{0pt}
  \setlength{\parsep}{0pt}
}{\end{enumerate}}
\usepackage{color, colortbl}

\usepackage[T1]{fontenc}

\usepackage[utf8]{inputenc}

\usepackage{microtype}

\usepackage{hyperref}

\makeatletter
\newcommand{\printfnsymbol}[1]{%
  \textsuperscript{\@fnsymbol{#1}}%
}
\makeatother

\let\SUP\textsuperscript

\title{Does Summary Evaluation Survive Translation to Other Languages?}

\author{Spencer Braun\SUP{1,$\ddagger$}, Oleg Vasilyev\SUP{1,$\ddagger$}, Neslihan Iskender\SUP{2,$\dagger$}, 
John Bohannon\SUP{1,$\ddagger$}\\
\SUP{1}Primer Technologies Inc., San Francisco, California\\
\SUP{2}Technische Universität Berlin, Quality and Usability Lab\\
\SUP{$\ddagger$}\texttt{spencer.braun,oleg,john}\texttt{@primer.ai} \\
\SUP{$\dagger$}\texttt{neslihan.iskender}\texttt{@tu-berlin.de}
}

\begin{document}
\maketitle
\begin{abstract}

The creation of a quality summarization dataset is an expensive, time-consuming effort, requiring the production and evaluation of summaries by both trained humans and machines. If such effort is made in one language, it would be beneficial to be able to use it in other languages without repeating human annotations. To investigate how much we can trust machine translation of such a dataset, we translate the English dataset \textit{SummEval} to seven languages and compare performance across automatic evaluation measures. We explore equivalence testing as the appropriate statistical paradigm for evaluating correlations between human and automated scoring of summaries. While we find some potential for dataset reuse in languages similar to the source, most summary evaluation methods are not found to be statistically equivalent across translations.

\end{abstract}

\section{Introduction}
A large summarization dataset includes thousands of texts and human-written summaries (for example, CNN/Daily Mail \cite{Karl2015Teaching}). In order to make it applicable for wider research, it may also contain machine-generated summaries by many models, accompanied by human and machine evaluations of the quality of the generated summaries \cite{Alexander2020SummEval}. The human annotation alone is a complicated effort, requiring careful planning and setup \cite{Iskender2020Reliability, Wojciech2020Evaluating}. As summarization resources grow for English-language models, it becomes increasingly important to consider whether we can repurpose these datasets for use in other languages as well.

In this paper, we focus on answering this question with respect to automated summarization evaluation using the SummEval dataset \cite{Alexander2020SummEval}, the largest corpus of English-language human annotated text summaries we could find. We translate this dataset from English to seven languages and evaluate the correlations between automated summary evaluation measures and human annotations. Using equivalence tests, we show that some aspects of summary quality ranking are preserved under translation for languages with similar alphabets and grammars to English. Additionally, we examine the behavior of automated measures across languages to compare the relative capabilities of machine translation versus summary evaluation. While we find some reasons for optimism about the potential for dataset reuse, our work clearly demonstrates that more research is needed to make translated datasets useful for a diverse set of languages. 


\section{Data}
We focus our analysis on the portion of SummEval\footnote{https://github.com/Yale-LILY/SummEval} that includes human annotations. It consists of 100 texts, each accompanied by 11 human-written reference summaries and 17 machine-generated summaries produced by different models. Each machine-generated summary is annotated by three experts and five crowd workers using a 5-point scale for four quality measures: coherence, consistency, fluency, and relevance. For simplicity, we create a composite rating by averaging the expert scores for each quality of a given text-summary pair. 

We translate all 100 source texts, 1100 human reference summaries, and 1700 machine-generated summaries into seven languages, French, German, Italian, Spanish, Afrikaans, Hindi, and Russian, using translation models trained and uploaded to the Hugging Face Model Hub by Helsinki-NLP\footnote{https://huggingface.co/Helsinki-NLP} and accessed via the transformers library \cite{Wolf2020Transformers} (see Appendix \ref{sec:models_used}). We used the same original human annotations provided by the SummEval dataset as annotations in each of the seven translated languages.

\begin{figure*}[ht!]
 \centering
    
    \begin{subfigure}[b]{0.49\textwidth}
      \includegraphics[width=1\textwidth]{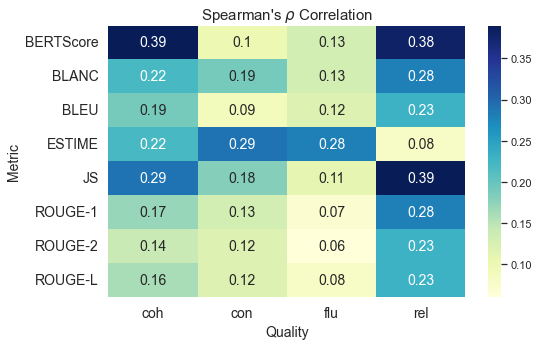}
    \end{subfigure}%
     \begin{subfigure}[b]{0.49\textwidth}
         \includegraphics[width=1\textwidth]{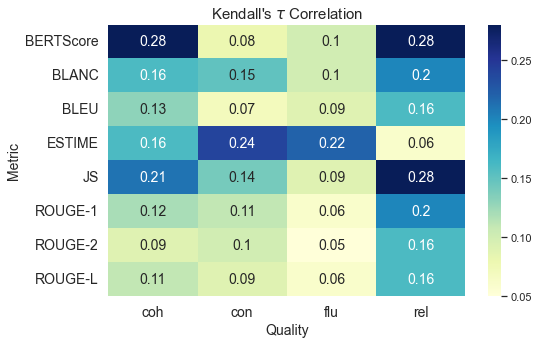}
    \end{subfigure}
 \caption{Spearman's $\rho$ and Kendall's $\tau$ correlations of expert human scores (coherence, consistency, fluency, relevance) with automated evaluation measures for the original English summaries. Note: JS (Jensen-Shannon) and ESTIME correlations are negated.}
 \label{tab:english_measures_human}
 \begin{subfigure}[b]{0.49\textwidth}
        \includegraphics[width=1\textwidth]{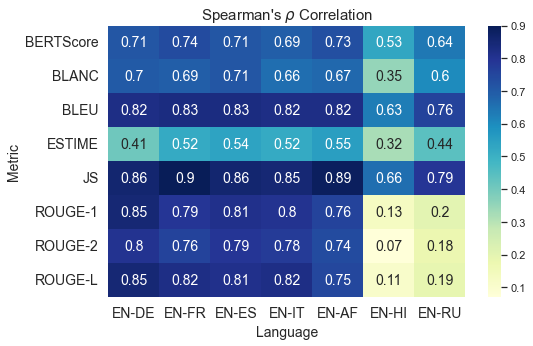}
        \vspace{0cm}
    \end{subfigure}%
    \begin{subfigure}[b]{0.49\textwidth}
        \includegraphics[width=1\textwidth]{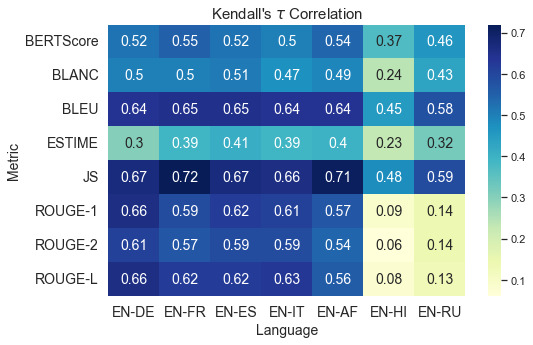}
        \vspace{0cm}
    \end{subfigure}
    \caption{Spearman's $\rho$ and Kendall's $\tau$ correlations between automated evaluation measures in English and in translated languages German (DE), French (FR), Spanish (ES), Italian (IT), Afrikaans (AF), Hindi (HI), and Russian (RU).}
    \label{tab:automatic-correlation}
\end{figure*}

In each language version of the dataset, we score machine-generated summaries with a few common or promising automated evaluation measures that could be applied to all eight languages. 
We calculate the following truly automated (not needing human written reference summaries) measures: Jensen-Shannon \cite{Annie2009Automatically}, ESTIME \cite{Oleg2020ESTIME}\footnote{https://github.com/PrimerAI/blanc/estime} and BLANC \cite{Oleg2020BLANC}\footnote{https://github.com/PrimerAI/blanc} (see Appendix \ref{sec:models_used}).
We also calculate the following reference-based automatic evaluation measures: BLEU \cite{Papineni2002BLEU}, BERTScore-F1\footnote{https://github.com/Tiiiger/bert\_score} \cite{Tianyi2020BERTScore}, and ROUGE \cite{Lin2004ROUGE} as ROUGE-1,2,L\footnote{https://github.com/google-research/google-research/tree/master/rouge}. 

We calculate correlations between automated evaluation measures in each language and the human annotations on the original English dataset. We seek to answer whether these correlations are reasonably independent of the language. In other words, can we rely on such correlations to provide consistent judgement of evaluation measures in other languages.

\section{Evaluation Correlations}
\label{sec:correlations}

\subsection{Correlations between Automated Measures and Expert Scores}
\label{sec:vs}

It has become standard in the summarization literature to judge the performance of an automated measure by the correlation of its scores with human evaluation of summaries (e.g. \cite{Tianyi2020BERTScore}, \cite{Deutsch2021stats}). Figure \ref{tab:english_measures_human} shows Spearman's $\rho$ and Kendall's $\tau$ correlation coefficients between the expert human evaluations and the automated measures run on the English summaries found in the SummEval dataset. 

The correlations are consistently weak, indicating that the measures rely on different features than human evaluations of a summary. ESTIME, BERTScore, and Jensen-Shannon all demonstrate somewhat higher correlations in at least some measures of quality, perhaps reflecting a more nuanced approach to summary scoring.

Automated evaluation of summarization models is still an evolving field. While most measures disagree with human judgment often, they are still widely used as points of comparison across model outputs. Therefore, it remains highly relevant to determine whether translation preserves the judgments rendered by the automated measures.

\subsection{Correlations for Automated Measures Across Languages}
\label{sec:using}

We may consider an evaluation measure to be useful under translation if the scores it assigns to summaries are consistent across languages, perhaps in absolute value but at least in the rank ordering of summaries. Therefore such a measure would exhibit high correlation between its values on English summaries and those for the summaries translated to other languages. Figure \ref{tab:automatic-correlation} shows Spearman's $\rho$ and Kendall's $\tau$ correlation coefficients between the automated measures run on the English corpus and each translated corpus. 

\begin{figure*}[t!]
 \centering
 \begin{subfigure}[b]{0.49\textwidth}
        \includegraphics[width=1\textwidth]{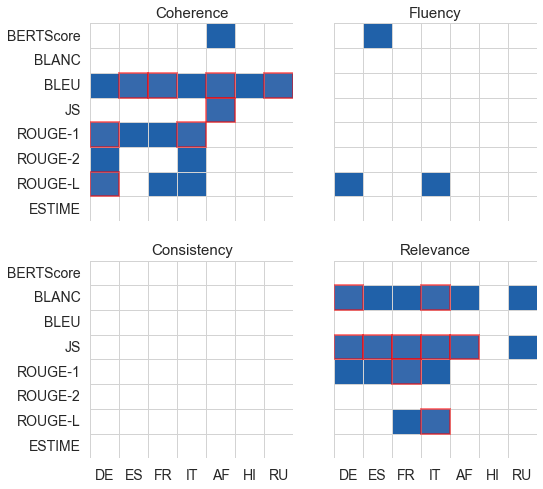}
        \vspace{0cm}
        \caption{TOST}
    \end{subfigure}%
    \begin{subfigure}[b]{0.49\textwidth}
        \includegraphics[width=1\textwidth]{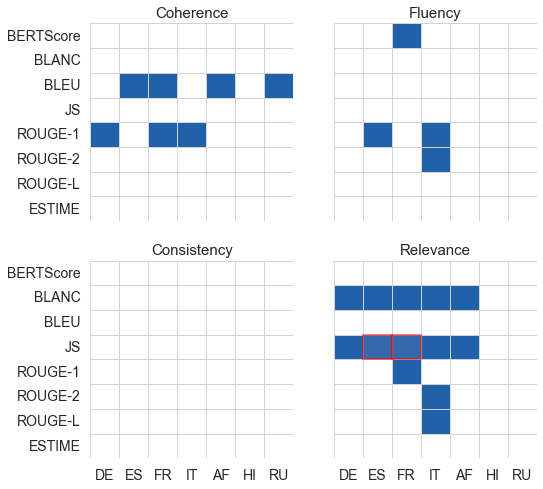}
        \vspace{0cm}
        \caption{Anderson-Hauck}
    \end{subfigure}
    \caption{Results of tests of equivalence for each automated measure, language, and quality measure (coherence, consistency, fluency, relevance). Blue squares indicate p-value $\leq 0.05$ while red highlights indicate the result remained significant after applying Benjamini-Yekutieli correction for FDR control. \textit{Left}: Results for TOST with standard deviation margin of equivalence. \textit{Right}: Results for Anderson-Hauck test with standard deviation margin of equivalence.}
    \label{fig:std_box}
\end{figure*}

For a given measure, the correlations across languages are generally much stronger than those between automated measures and human evaluations in English seen in Figure \ref{tab:english_measures_human}. For languages with the strongest correlations to the English measures, this result provides some promise that translation might introduce minimal additional noise, meaning the evaluation measure provides consistent signal across languages.

The reference-based measures generally show stronger correlations ($\rho > 0.6$, $\tau > 0.5$) between English and German, French, Spanish, Italian, and Afrikaans translations. For Russian and Hindi, they show weaker correlations, drastically so for ROUGE measures. Among the reference-free measures, Jensen-Shannon and BLANC demonstrate similar patterns of performance. These results at least suggest that measures may prove useful when translating datasets to languages with similar origins (here Italic or Germanic languages). However, ESTIME shows weak correlations across languages with a smaller drop in correlation between Western European derived languages and Hindi and Russian.

\section{Significance Tests}
\label{sec:sig}

Given the promising results in Section \ref{sec:correlations}, we seek to test whether correlations between an automated measure and the original expert scores are statistically invariant when run on the English and translated summaries. Since human evaluations are split into four qualities - coherence, consistency, fluency, relevance - we consider correlations separately along each measure. For example, we look to answer whether the correlation between English BLANC scores and English expert scores for relevance is equivalent to the correlation between German BLANC scores and English expert scores for relevance. We consider this a natural test of an automated measure's utility after translation, as we hope measures will reflect human judgment in a consistent and predictable manner across languages. 

Since we are interested in demonstrating a lack of statistical difference between two correlations, $\rho_1$ and $\rho_2$, we cannot use a typical hypothesis test with null hypothesis $H_0: \rho_1 = \rho_2$. Such a test would only suggest equivalence by failing to reject the null hypothesis, which could simply occur due to a lack of statistical power.

Instead, we turn to equivalence tests, a paradigm which effectively reverses null and alternative hypotheses, ie. $H_0: \rho_1 \neq \rho_2$. We explore two such tests, Two One-Sided Tests (TOST) and Anderson-Hauck tests, and call for additional research to standardize their use for summarization evaluation.

\subsection{Two One-Sided Tests (TOST)}
\label{sec:tost}

In the TOST procedure \cite{Schuirmann2005ACO}, we must set a margin of equivalence, $\Delta_E$, within which we consider two test statistics to be equivalent. Then for two correlations, $\rho_1$ and $\rho_2$, we have null and alternative hypotheses $H_0: \, \rho_1 - \rho_2 < -\Delta_E \text{ or } \rho_1 - \rho_2 > \Delta_E$ and $H_1: -\Delta_E < \rho_1 - \rho_2 < \Delta_E$. While in a field like medicine, the margin might be well defined by a chemical process, we lack a strong prior for choosing a relevant margin. We explore several options and consider the sensitivity of p-values to our choices when evaluating the validity of the tests' conclusions.

The Kendall rank correlation differences considered do not follow a normal distribution, and we use bootstrap resampling \cite{EfroTibs93} to generate an empirical distribution. For a given translation language, automated evaluation measure, and quality measure, we sample across (text, summary, and reference summary) tuples. (Note for reference-based summaries - BERTScore, BLEU, and ROUGE - a more complete bootstrap procedure would account for the stochasticity present in the choice of reference summaries themselves. We provide an illustrative example in Appendix \ref{sec:bootstrap_refs}.)

While permutation-based tests have been shown to have higher power in summarization evaluation than bootstrap resampling \cite{Deutsch2021stats}, permutation tests assume null hypothesis $H_0: \rho_1= \rho_2$ and are not simply adapted to our case. We apply a multiple testing correction to the p-values calculated due to the large number of tests considered. We use the Benjamini-Yekutieli procedure \cite{benjamini2001by} to account for dependence among correlation measures and control the false discovery rate (FDR) at level $\alpha=0.05$.

Equivalence margins considered include a flat margin of 0.05, the standard deviations for correlations between individual experts and an automated measure, and the maximum difference for correlations between individual experts and an automated measure. Under the constant 0.05 margin, 56\% of correlations are equivalent before correction and 31\% after. Under the standard deviation margin, 17\% of tests are equivalent before and 8\% after correction. Finally, under the max difference margin, 42\% of correlations are equivalent before corrections and 28\% after.

We present the full results of the TOST procedure with a standard deviation margin in the left panel of Figure \ref{fig:std_box} and can note a few clear patterns.  First, as seen under the simple correlation analysis, the Italic and Germanic languages have a higher number of significant results than Hindi or Russian. We may still consider using translated summarization datasets from English to languages considered "close." However, there are few significant results in the fluency or consistency qualities. Therefore the automated measures may only be useful under translation along specific dimensions of quality. Looking at the correlations in English between automated measures and expert judgments in Figure \ref{tab:english_measures_human}, fluency and consistency also tend to have much lower correlations than coherence and relevance.

\begin{figure}[h]
\centering
\includegraphics[scale=0.50]{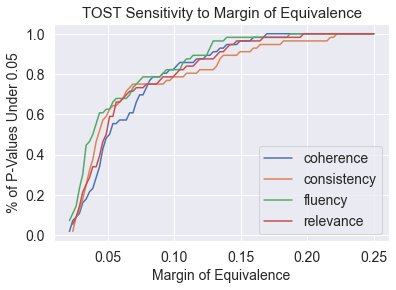}
\caption{Sensitivity of TOST to the margin of equivalence. Small changes in the margin can result in a large change in the percent of tests with significant p-values.}
\label{fig:tost1}
\end{figure}

Clearly the choice of equivalence margin has a consequential impact on results. Figure \ref{fig:tost1} shows how the number of significant p-values changes in response to an increasing margin of equivalence. Given the apparent sensitivity to changes in the margin, further research is warranted into how the performance of translation and summarization systems relates to the correlations measured here. 

\subsection{Anderson-Hauck Tests}
\label{sec:ah}

While TOST provides a non-parametric route towards equivalence testing, we consider an additional parametric test that may improve statistical power. The Anderson-Hauck test is an equivalence testing procedure for dependent correlation coefficients which uses an approximate non-central t-distribution to calculate p-values \cite{andersonhauck1983}. Prior comparisons with TOST demonstrated that Anderson-Hauck can trade some additional Type-I error for higher power \cite{Counsell2015equivalence}. 

We consider the same margins of equivalence and apply Benjamini-Yekutieli for FDR control at level $\alpha=0.05$. A similar pattern emerges when considering results under different margins, and under the standard deviation margin we reject the null hypothesis in under 1\% of tests. The full results for the standard deviation margin are presented in the right panel of Figure \ref{fig:std_box}.

The pattern of equivalence is largely the same as that found under TOST but with greater sparsity of significant results. Ultimately while the tests hint towards the ability to reuse summarization datasets in similar languages to English, we are only able to detect equivalence in a minority of cases.

\section{Discussion}
\label{sec:discussion}

The correlation results presented significant differences among automated summarization measures and their relationships to the four quality measures. We seek to build an intuition for these findings and make use of qualitative exploration and round-trip translation to ground our understanding.

\begin{figure}[h]
\centering
\includegraphics[width=\linewidth]{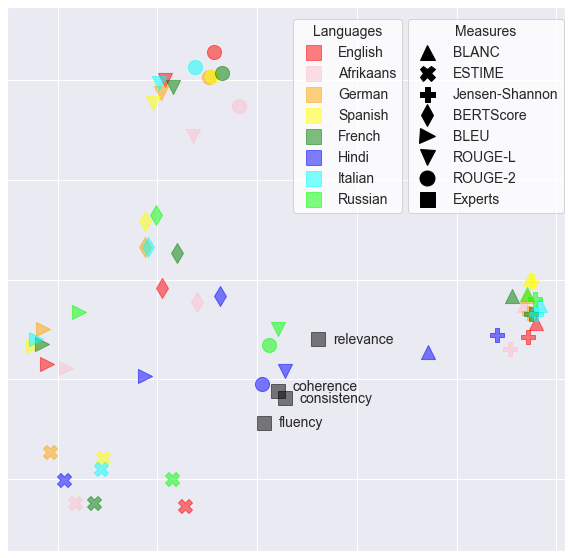}
\caption{PCA plot of summary quality scores. All scores were transformed to ranks before PCA, to reduce subjectivity of the respective scales. Note the human expert scores in black squares exist for the English dataset only.}
\label{fig:PCA}
\end{figure}

\subsection{Qualitative Analysis}
\label{sec:pca}
We can review the scores for the 1700 summaries in reduced dimensions using principal components analysis (PCA). Figure \ref{fig:PCA} shows each 1700-dimensional vector projected onto the first two principal components, which collectively explain 38.5\% of the variance. There are four vectors of human expert scores, corresponding to the quality measures coherence, consistency, fluency, and relevance, averaged over the three individual experts. Each automated measure (for example, ROUGE-2) produced eight 1700-dimensional vectors, one for each language. 

PCA can be used to disentangle the sources of divergence among evaluation measures under translation. The plot helps highlight the relative strength of translation over the summarization evaluation methods themselves. If machine translation added significant noise to the summaries, we would expect the relative position of language-specific scores in Figure \ref{fig:PCA} to be inconsistent across evaluation measures. Instead, we generally observe tight clusters for each evaluation measure with shared relative positions among the languages (at least when ignoring Hindi and Russian). This pattern suggests exploring the "stability" of evaluation measures undergoing translation - are we able to recover the original ranking of automated scores when we translate back to English? 

We note a few curious observations from Figure \ref{fig:PCA} in Appendix \ref{sec:observatios_pca}.

\subsection{Measure-Measure Comparisons}
\label{sec:roundtrip}

While our statistical tests focus on the absolute correlation between automated and human scores, the PCA suggests considering translation as a shift in geometric space. We can consider the automated measures relative to one another: if one measure correlates better than another with human scores in the original (English) dataset, would it still be better in a translated (non-English) dataset? Additionally, we can return the dataset back to English to get a sense of the distortion introduced by the translation process. 

To estimate the consistency with which one measure dominates another, we turn to bootstrap resampling of the summary evaluations. We select 10,000 bootstrap samples from the 1700 text-summary-references tuples. Let $P$ represent the fraction of samples in which one measure is better than another for a given measure-measure pair; we consider a pair resolved if one measure outperforms another in at least $97.5\%$ of all the resamplings, ie. $P \geq 0.975$ in the original English dataset. Using Kendall rank correlations, the number of resolved measure-measure pairs is 61\% for coherence, 56\% for consistency, 42\% for fluency and 64\% for relevance. With a baseline reading of how stable the measure rankings are in English, we can ask what happens with these resolved pairs when the dataset is translated. 

\begin{figure}[h]
\centering
\includegraphics[scale=0.52]{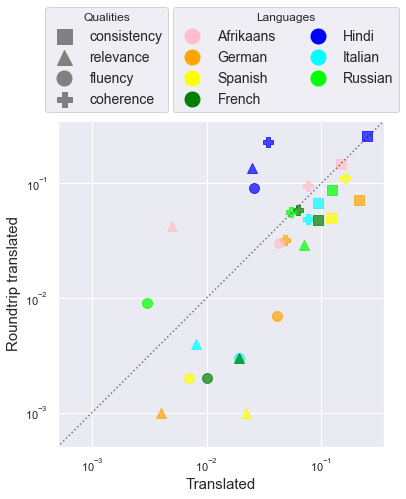}
\caption{Result of bootstrapping: average shift in probability $P$ of one measure being better than another, when the evaluation data are translated to another language (x-axis) and then translated back to English (y-axis). The average is taken over all measure-measure pairs that had $P \geq 0.975$ in English.}
\label{fig:Bootstrap_pairs}
\end{figure}

Along its x-axis, Figure \ref{fig:Bootstrap_pairs} shows how much on average the fraction $P$ changes (increases or decreases) after translation for resolved measure-measure pairs, where the average is over a given language and quality measure.  
For most languages and qualities the shift of $P$ is less than 0.1, the largest is 0.25 (consistency, Hindi). Many resolved measure-measure pairs become unresolved after translation, though no shift is drastic enough to reverse which measure ranks higher in a majority of samples (i.e. crossing $P=0.5$). Figure \ref{fig:Bootstrap_pairs} suggests that in most cases our conclusion about comparing two measures will not change with translation.


A round-trip translation returns each summary to its source language, allowing us to more effectively isolate the effect of translation quality on the consistency of automated measures. The dashed line $y=x$ seen in Figure \ref{fig:Bootstrap_pairs} represents points where the round-trip translation causes an equally-sized shift as the forward translation. We note that the observed shifts are mostly under the diagonal - the shifts caused by translation are to some degree reversed when we return to English. 

While the shifts for round-trip translations are on average smaller than for one-way, they demonstrate that translation is far from perfect and introduces enough noise to be detected by the summarization evaluation measures.
Notably, the points above the diagonal come from Hindi, Russian and Afrikaans round-trip translation. This confirms our intuition that a translation to languages more distant from English is more risky for the survival of the summary evaluation. We hope further research may reveal a way to use the round-trip translation for the criteria of survival.



\section{Conclusion}

In this paper, we explored how well automated evaluation of summarization remain consistent on texts and summaries translated to other languages. We focused on the SummEval dataset and considered its translation to French, German, Italian, Spanish, Afrikaans, Hindi, and Russian. 

Additionally, we explored tests of equivalence as a statistical method of evaluating the performance of automated measures under translation. We found that translation can preserve correlations of evaluation metrics with the English human scores for coherence or relevance but could not conclude the same for fluency or consistency.

We call for additional research into summary evaluation metrics that can survive translation, as it offers a relatively simple path towards extending NLP capabilities for lower resource languages. Future work could identify how changes in the margin of equivalence equate to deterioration of model performance. Additionally, this line of research could be extended to a larger selection of languages and automated evaluation measures.


\bibliography{anthology,custom}

\begin{thebibliography}{17}
\expandafter\ifx\csname natexlab\endcsname\relax\def\natexlab#1{#1}\fi

\bibitem[{Anderson and Hauck(1983)}]{andersonhauck1983}
Sharon Anderson and Walter~W. Hauck. 1983.
\newblock \href {https://doi.org/10.1080/03610928308828634} {A new procedure
  for testing equivalence in comparative bioavailability and other clinical
  trials}.
\newblock \emph{Communications in Statistics - Theory and Methods},
  12(23):2663--2692.

\bibitem[{Benjamini and Yekutieli(2001)}]{benjamini2001by}
Yoav Benjamini and Daniel Yekutieli. 2001.
\newblock \href {https://doi.org/10.1214/aos/1013699998} {{The control of the
  false discovery rate in multiple testing under dependency}}.
\newblock \emph{The Annals of Statistics}, 29(4):1165 -- 1188.

\bibitem[{Counsell and Cribbie(2015)}]{Counsell2015equivalence}
Alyssa Counsell and Robert~A. Cribbie. 2015.
\newblock \href {https://doi.org/https://doi.org/10.1111/bmsp.12045}
  {Equivalence tests for comparing correlation and regression coefficients}.
\newblock \emph{British Journal of Mathematical and Statistical Psychology},
  68(2):292--309.

\bibitem[{Deutsch et~al.(2021)Deutsch, Dror, and Roth}]{Deutsch2021stats}
Daniel Deutsch, Rotem Dror, and Dan Roth. 2021.
\newblock \href {https://doi.org/10.1162/tacl_a_00417} {{A Statistical Analysis
  of Summarization Evaluation Metrics Using Resampling Methods}}.
\newblock \emph{Transactions of the Association for Computational Linguistics},
  9:1132--1146.

\bibitem[{Efron and Tibshirani(1993)}]{EfroTibs93}
Bradley Efron and Robert~J. Tibshirani. 1993.
\newblock \emph{An Introduction to the Bootstrap}.
\newblock Number~57 in Monographs on Statistics and Applied Probability.
  Chapman \& Hall/CRC, Boca Raton, Florida, USA.

\bibitem[{Fabbri et~al.(2021)Fabbri, Kryściński, McCann, Xiong, Socher, and
  Radev}]{Alexander2020SummEval}
Alexander~R. Fabbri, Wojciech Kryściński, Bryan McCann, Caiming Xiong,
  Richard Socher, and Dragomir Radev. 2021.
\newblock \href {https://doi.org/10.1162/tacl_a_00373} {{SummEval}:
  Re-evaluating summarization evaluation.}
\newblock \emph{Transactions of the Association for Computational Linguistics},
  9:391--409.

\bibitem[{Hermann et~al.(2015)Hermann, Kočiský, Grefenstette, Espeholt, Kay,
  Suleyman, and Blunsom}]{Karl2015Teaching}
Karl~Moritz Hermann, Tomáš Kočiský, Edward Grefenstette, Lasse Espeholt,
  Will Kay, Mustafa Suleyman, and Phil Blunsom. 2015.
\newblock \href
  {https://proceedings.neurips.cc/paper/2015/file/afdec7005cc9f14302cd0474fd0f3c96-Paper.pdf}
  {Teaching machines to read and comprehend.}
\newblock In \emph{Advances in Neural Information Processing Systems 28}, pages
  1693--1701. Curran Associates, Inc.

\bibitem[{Iskender et~al.(2021)Iskender, Polzehl, and
  Möller}]{Iskender2020Reliability}
Neslihan Iskender, Tim Polzehl, and Sebastian Möller. 2021.
\newblock \href {https://aclanthology.org/2021.humeval-1.10} {Reliability of
  human evaluation for text summarization: Lessons learned and challenges
  ahead.}
\newblock In \emph{Proceedings of the Workshop on Human Evaluation of NLP
  Systems (HumEval)}, pages 86--96. Association for Computational Linguistics
  (2021).

\bibitem[{Kryscinski et~al.(2020)Kryscinski, McCann, Xiong, and
  Socher}]{Wojciech2020Evaluating}
Wojciech Kryscinski, Bryan McCann, Caiming Xiong, and Richard Socher. 2020.
\newblock \href {https://aclanthology.org/2020.emnlp-main.750} {Evaluating the
  factual consistency of abstractive text summarization.}
\newblock In \emph{Proceedings of the 2020 Conference on Empirical Methods in
  Natural Language Processing}, pages 9332--9346. Association for Computational
  Linguistics.

\bibitem[{Lin(2004)}]{Lin2004ROUGE}
Chin-Yew Lin. 2004.
\newblock \href {https://aclanthology.org/W04-1013} {{ROUGE}: A package for
  automatic evaluation of summaries.}
\newblock In \emph{Proceedings of Workshop on Text Summarization Branches Out},
  pages 74--81. Association for Computational Linguistics.

\bibitem[{Louis and Nenkova(2009)}]{Annie2009Automatically}
Annie Louis and Ani Nenkova. 2009.
\newblock \href {https://aclanthology.org/D09-1032} {Automatically evaluating
  content selection in summarization without human models.}
\newblock In \emph{Proceedings of the 2009 Conference on Empirical Methods in
  Natural Language Processing}, pages 306--314. Association for Computational
  Linguistics.

\bibitem[{Papineni et~al.(2002)Papineni, Roukos, Ward, and
  Zhu}]{Papineni2002BLEU}
Kishore Papineni, Salim Roukos, Todd Ward, and Wei-Jing Zhu. 2002.
\newblock \href {https://aclanthology.org/P02-1040} {{BLEU}: a method for
  automatic evaluation of machine translation.}
\newblock In \emph{Proceedings of the 40th Annual Meeting of the Association
  for Computational Linguistics (ACL)}, pages 311--318, Philadelphia.
  Association for Computational Linguistics.

\bibitem[{Schuirmann(1987)}]{Schuirmann2005ACO}
Donald~J. Schuirmann. 1987.
\newblock A comparison of the two one-sided tests procedure and the power
  approach for assessing the equivalence of average bioavailability.
\newblock \emph{Journal of Pharmacokinetics and Biopharmaceutics}, 15:657--680.

\bibitem[{Vasilyev and Bohannon(2021)}]{Oleg2020ESTIME}
Oleg Vasilyev and John Bohannon. 2021.
\newblock \href {https://aclanthology.org/2021.eval4nlp-1.10} {Estime:
  Estimation of summary-to-text inconsistency by mismatched embeddings.}
\newblock In \emph{Proceedings of the 2nd Workshop on Evaluation and Comparison
  of NLP Systems}, pages 94--103. Association for Computational Linguistics.

\bibitem[{Vasilyev et~al.(2020)Vasilyev, Dharnidharka, and
  Bohannon}]{Oleg2020BLANC}
Oleg Vasilyev, Vedant Dharnidharka, and John Bohannon. 2020.
\newblock \href {https://aclanthology.org/2020.eval4nlp-1.2} {Fill in the
  {BLANC}: Human-free quality estimation of document summaries.}
\newblock In \emph{Proceedings of the First Workshop on Evaluation and
  Comparison of NLP Systems}, pages 11--20. Association for Computational
  Linguistics.

\bibitem[{Wolf et~al.(2020)Wolf, Debut, Sanh, Chaumond, Delangue, Moi, Cistac,
  Rault, Louf, Funtowicz, Davison, Shleifer, von Platen, Ma, Jernite, Plu, Xu,
  Scao, Gugger, Drame, Lhoest, and Rush}]{Wolf2020Transformers}
Thomas Wolf, Lysandre Debut, Victor Sanh, Julien Chaumond, Clement Delangue,
  Anthony Moi, Pierric Cistac, Tim Rault, Remi Louf, Morgan Funtowicz, Joe
  Davison, Sam Shleifer, Patrick von Platen, Clara Ma, Yacine Jernite, Julien
  Plu, Canwen Xu, Teven~Le Scao, Sylvain Gugger, Mariama Drame, Quentin Lhoest,
  and Alexander Rush. 2020.
\newblock \href {https://aclanthology.org/2020.emnlp-demos.6} {Transformers:
  State-of-the-art natural language processing.}
\newblock In \emph{Proceedings of the 2020 Conference on Empirical Methods in
  Natural Language Processing: System Demonstrations}, pages 38--45.
  Association for Computational Linguistics (2020).

\bibitem[{Zhang et~al.(2020)Zhang, Kishore, Wu, Weinberger, and
  Artzi}]{Tianyi2020BERTScore}
Tianyi Zhang, Varsha Kishore, Felix Wu, Kilian~Q. Weinberger, and Yoav Artzi.
  2020.
\newblock \href {http://arxiv.org/abs/1904.09675v3} {{BERTScore}: Evaluating
  text generation with bert.}
\newblock \emph{arXiv}, arXiv:1904.09675v3.

\end{thebibliography}
\bibliographystyle{acl_natbib}

\appendix

\section{Model Details for Translation and Evaluation Measures}\label{sec:models_used}
All the transformer models we used are from the transformers library \cite{Wolf2020Transformers}.
The models used for translation\footnote{https://huggingface.co/Helsinki-NLP} are all "opus-mt-L1-L2", where one of L1 or L2 is 'en' (English), and the other is one of the languages 'af', 'de', 'es', 'fr', 'hi', 'it', 'ru'.

We used 'bert-base-multilingual-cased' as the underlying model for BLANC and ESTIME. BERTScore also relies on 'bert-base-multilingual-cased' for all languages except English, for which it uses the model 'roberta-large' \footnote{https://github.com/Tiiiger/bert\_score}. ESTIME embeddings were taken from the 10th transformer block layer, below the top layer for the base BERT (we followed ~\cite{Oleg2020ESTIME}, where it was shown that for the larger model 'bert-large-uncased-whole-word-masking' the 21st layer, also below the top, delivers the best performance).

\section{Observations from PCA}\label{sec:observatios_pca}
The locations of the measures in Figure \ref{fig:PCA} after translation largely remain close to the original English version, except Hindi and Russian points. The locations show interesting patterns. The reference-based measures, based on hard (ROUGE, BLEU) or soft (BERTScore) overlap of tokens between the summary and the human-written reference summaries, are in the same top left quadrant with respect to the human scores. The reference-free measures BLANC and Jensen-Shannon are on the opposite side. It is natural for both BLANC and Jensen-Shannon to be on the relevance side of the human scores: BLANC estimates how well a text can be reconstructed from its summary, and Jensen-Shannon considers the Kullback–Leibler divergence between the summary and the text. For ESTIME, however, as for a consistency-oriented measure, it makes sense to be on the consistency and fluency side of the human scores, rather than on the relevance side.

For most measures, the translated scores are often closer to the expert evaluations than the English scores. Strangely, it is especially true for Hindi and, in the case of ROUGE, for Russian. One possible explanation is that the translation simplifies the phrases and the choice of words, thus making it easier for some evaluation measures, at least along some dimensions. The pattern associated with ESTIME is distinct from other measures: the non-English scores for ESTIME are almost always further away from the human scores. This suggests that maybe ESTIME is sensitive enough to require a higher quality translation. We cannot blame the underlying multilingual model, because both BLANC and BERTScore use the same model (see Appendix \ref{sec:models_used}). 

\section{Bootstrap with Reference-Summaries}\label{sec:bootstrap_refs}
Throughout the paper we used bootstrapping with resampling of the (text, summary, references) tuples, where the references are the reference summaries needed by some measures (BERTScore, BLEU, ROUGE). For each text in SummEval \cite{Alexander2020SummEval}, there are 11 reference summaries, and a full bootstrap for the reference-based measures should also include a resampling of the reference summaries themselves. 

The impact of this added source of randomness can be seen by constructing confidence intervals for the estimated correlation between an evaluation measure and human scores. When we add resampling over reference summaries, confidence intervals widen and require more time and resources to compute. In Table \ref{tab:Widening_refsummaries} we illustrate the widening of the confidence interval on an example using BERTScore correlations with SummEval human expert scores (in the original English SummEval dataset). We ran 500K reference summaries resamplings, recomputing scores and correlations. The BERTScore is a peculiar and convenient case for bootstrap resampling of reference summaries, because the score is defined as a max score over the scores taken individually for each reference summary \cite{Tianyi2020BERTScore}.

\begin{table}[h]
\centering
\scalebox{0.75}{
\begin{tabular}{l|lll|lll}
 & \multicolumn{3}{c|}{\textbf{Kendall's $\tau$}} &
 \multicolumn{3}{c}{\textbf{Spearman's $\rho$}} \\ \toprule
 & \textbf{low} & \textbf{high} & \textbf{widen} & \textbf{low} & \textbf{high} & \textbf{widen} \\ \midrule
\textbf{coherence} & 0.245 & 0.307 & 0.011 & 0.345 & 0.428 & 0.015 \\
\textbf{consistency} & 0.041 & 0.117 & 0.002 & 0.052 & 0.148 & 0.003 \\
\textbf{fluency} & 0.062 & 0.135 & 0.004 & 0.080 & 0.175 & 0.006 \\
\textbf{relevance} & 0.246 & 0.310 & 0.026 & 0.338 & 0.424 & 0.035 \\
\bottomrule
\end{tabular}}
\caption{The columns 'low' and 'high' are the confidence boundaries from bootstrap without resampling reference summaries, for BERTScore correlations with expert human scores (coherence, consistency, fluency, relevance). The column 'widen' is the widening of the confidence interval as a result of adding the resampling of the reference summaries to the bootstrap resampling. Kendall's Tau correlation is Tau-c. The confidence boundaries are for 0.025 and 0.975 percentiles. The bootstrapping used 500K resamplings.}
\label{tab:Widening_refsummaries}
\end{table}

The low and high correlation values are given in the table for bootstrap without resampling of reference summaries, as corresponding to 0.025 and 0.975 percentiles of the distribution. The 'widen' column in the table shows how much the confidence interval ('high minus low') changed after including resampling of the 11 reference summaries into the bootstrapping. Some quality measures are especially affected by the change, with confidence intervals for Kendall correlation widening by 40\% for relevance and by 17\% for coherence (for Spearman's correlations, correspondingly, 42\% and 18\%). Notice that the relevance and coherence are exactly the qualities in which BERTScore is reported as a strong measure \cite{Oleg2020ESTIME}.

\end{document}